\documentclass[runningheads]{llncs}
\usepackage{graphicx}

\usepackage{tikz}
\usepackage{comment} 
\usepackage{amsmath,amssymb} 
\usepackage{color}

\usepackage{tabularx}
\usepackage{multirow}
\usepackage{booktabs}
\usepackage[pagebackref=true,breaklinks=true,letterpaper=true,colorlinks,bookmarks=false]{hyperref}
\usepackage{threeparttable}

\begin{document}
\pagestyle{headings}
\mainmatter
\def\ECCVSubNumber{4152}  

\title{Kernelized Memory Network for Video Object Segmentation} 

\titlerunning{Kernelized Memory Network for Video Object Segmentation}
%
\author{Hongje Seong \and 
Junhyuk Hyun \and
Euntai Kim\thanks{Corresponding author.}} 
%
\authorrunning{H. Seong et al.}
%
\institute{School of Electrical and Electronic Engineering, Yonsei University, Seoul, Korea \\
\email{\{hjseong,jhhyun,etkim\}@yonsei.ac.kr}
}
\maketitle

\begin{abstract}
Semi-supervised video object segmentation (VOS) is a task that involves predicting a target object in a video when the ground truth segmentation mask of the target object is given in the first frame. Recently, space-time memory networks (STM) have received significant attention as a promising solution for semi-supervised VOS. However, an important point is overlooked when applying STM to VOS. The solution (STM) is non-local, but the problem (VOS) is predominantly local. To solve the mismatch between STM and VOS, we propose a kernelized memory network (KMN). Before being trained on real videos, our KMN is pre-trained on static images, as in previous works. Unlike in previous works, we use the Hide-and-Seek strategy in pre-training to obtain the best possible results in handling occlusions and segment boundary extraction.
The proposed KMN surpasses the state-of-the-art on standard benchmarks by a significant margin (+5\% on DAVIS 2017 test-dev set).
In addition, the runtime of KMN is 0.12 seconds per frame on the DAVIS 2016 validation set, and the KMN rarely requires extra computation, when compared with STM.
\end{abstract}

\section{Introduction}
\label{s1}
Video object segmentation (VOS) is a task that involves tracking target objects at the pixel level in a video. It is one of the most challenging problems in computer vision. VOS can be divided into two categories: semi-supervised VOS and unsupervised VOS. In semi-supervised VOS, the ground truth (GT) segmentation mask is provided in the first frame, and the segmentation mask must be predicted for the subsequent frames. In unsupervised VOS, however, no GT segmentation mask is provided, and the task is to find and segment the salient object in the video. In this paper, we consider semi-supervised VOS.

\begin{figure}[t]
\centering
\includegraphics[width=0.7\linewidth]{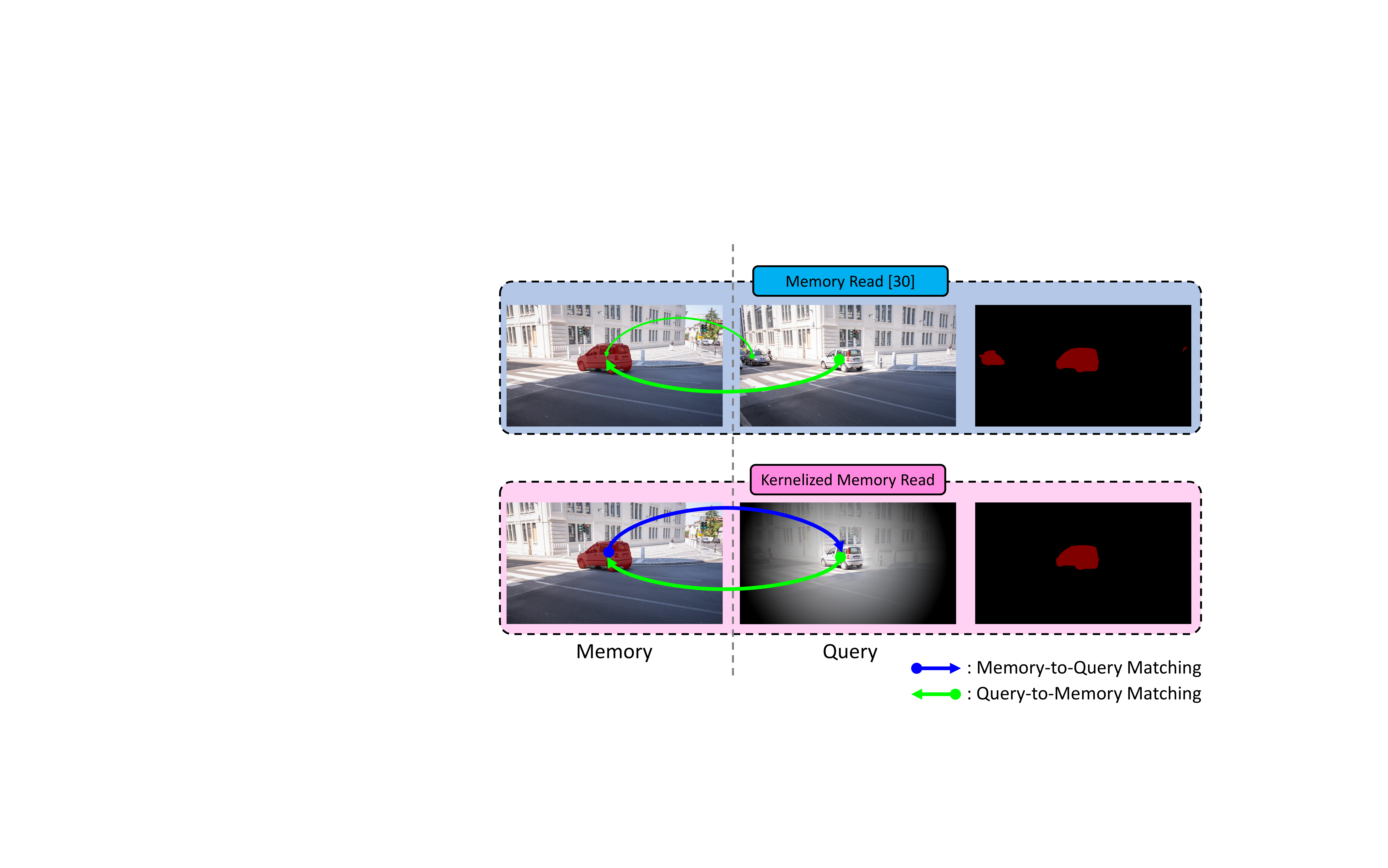}
\caption{
Illustration of KMN. In STM \cite{Oh_2019_ICCV}, two cars in the query frame are matched with a car in the memory frame owing to the non-local matching between the query and memory. The car in the middle is the correct match, while the car on the left is an incorrect match. In KMN, however, non-local matching between the query and memory is controlled by the Gaussian kernel. Only the car in the middle of the query frame is matched with the car in the memory.
}
\label{fig:visualize_GaussianKernel}
\end{figure}

Space-time memory networks (STM) \cite{Oh_2019_ICCV} have recently received significant attention as a promising solution for semi-supervised VOS. The basic idea behind the application of STM to VOS is to use the intermediate frames between the first frame and the current frame. In STM, the current frame is considered to be the query frame for which the target is to be predicted, whereas the past (already predicted) frames are used as memory frames. This approach, however, overlooks an important point. The solution (STM) is non-local, but the problem (VOS) is predominantly local, as illustrated in Fig. \ref{fig:visualize_GaussianKernel}. Specifically, STM is based on non-local matching between the query frame and memory frames. However, in VOS, the target object in the query frame usually appears in the local neighborhood of the target’s appearance in the memory frames. To solve the problem arising from the use of STM for VOS, we propose a kernelized memory network (KMN). In KMN, the Gaussian kernel is employed to reduce the degree of non-localization of the STM and improve the effectiveness of the memory network for VOS.

Before being trained on real videos, our KMN is pre-trained on static images, as in some previous works. In particular, multiple frames based on a random affine transform were used in \cite{wug2018fast,Oh_2019_ICCV}. Unlike the training process in the previous works, however, we employ a Hide-and-Seek strategy during pre-training to obtain the best possible results in handling occlusions and segment boundary extraction. The Hide-and-Seek strategy \cite{singh2017hide} was initially developed for weakly supervised object localization, but we used it to pre-train the KMN. This provides two key benefits. First, Hide-and-Seek achieves segmentation results that are considerably robust to occlusion. To the best of our knowledge, this is the first time that Hide-and-Seek has been applied to VOS in order to make the predictions robust to occlusion. Second, Hide-and-Seek is used to refine the boundary of the object segment. Because most of the ground truths in segmentation datasets contain unclear and incorrect boundaries, it is fairly challenging to predict accurate boundaries in VOS. The boundaries created by Hide-and-Seek, however, are clear and accurate. Hide-and-seek appears to provide instructive supervision for clear and precise cuts for objects, as shown in Fig. \ref{fig:hide-and-seek}. We conduct experiments on DAVIS 2016, DAVIS 2017, and Youtube-VOS 2018 and significantly outperform all previous methods, even compared with online-learning approaches.

The contributions of this paper can be summarized as follows. First, KMN is developed to reduce the non-locality of the STM and make the memory network more effective for VOS. Second, Hide-and-Seek is used to pre-train the KMN on static images.

\section{Related Work}
\label{s2}

\subsubsection{Semi-supervised video object segmentation}\cite{perazzi2016benchmark,pont20172017,xu2018youtube} is a task involving prediction of the target objects in all frames of a video sequence where information of the target objects is provided in the first frame. Because the object mask for the first frame of the video is given at the test time, many previous studies \cite{shin2017pixel,cheng2017segflow,perazzi2017learning,caelles2017one,hu2017maskrnn,ci2018video,bao2018cnn,luiten2018premvos,maninis2018video,voigtlaender2017online,li2018video,Wang_2019_ICCV} fine-tuned their networks on the given mask. This is known as the online-learning strategy. Online-learning methods can provide accurate prediction results, but require considerable time for inference and finding the best hyper-parameters of the model for each sequence. Offline-learning methods \cite{marki2016bilateral,jampani2017video,yang2018efficient,chen2018blazingly,voigtlaender2019feelvos,wug2018fast,cheng2018fast,Zhang_2019_ICCV,Oh_2019_ICCV} use a fixed parameter set trained on the whole training sequence. Therefore, they can have a fast run time, while achieving comparable accuracy. Our proposed method follows the offline approach.

\subsubsection{Memory networks}\cite{sukhbaatar2015end} use the query, \textbf{key}, and \textbf{value} (QKV) concept. The QKV concept is often used when the target information of the current input exists at the other inputs. In this case, memory networks set the current input and the other inputs as the query and memory, respectively. The \textbf{key} and \textbf{value} are extracted from memory, and the correlation map of the query and memory is generated through a non-local matching operation of the query and \textbf{key} feature. Then, the weighted average \textbf{value} based on the correlation map is retrieved. The QKV concept is widely used in a variety of tasks, including natural language processing \cite{vaswani2017attention,miller2016key,kumar2016ask}, image processing \cite{parmar2018image,Zhu_2019_ICCV}, and video recognition \cite{wang2018non,girdhar2019video,seong2019video}. In VOS, STM \cite{Oh_2019_ICCV} has achieved significant success by repurposing the concept of the QKV. However, applications in STM tend to overlook an important feature of VOS, leading to a limitation that will be addressed in this paper.

\subsubsection{Kernel soft argmax}\cite{lee2019sfnet} uses Gaussian kernels on the correlation map to create a gradient propagable argmax function for semantic correspondence. The semantic correspondence task requires only a single matching flow from a source image to a target image for each given source point. However, applying a discrete argmax function on the correlation map makes the network untrainable. To solve this problem, kernel soft argmax applies Gaussian kernels on the correlation map and then averages the correlation scores. Our work is inspired by the kernel soft argmax, but its application and objective are completely different. The kernel soft argmax applies Gaussian kernels to the results of the searching flow (\textit{i.e.}, memory frame) to serve as a gradient propagable argmax function, whereas we applied Gaussian kernels on the opposite side (\textit{i.e.}, query frame) to solve the case as shown in Fig. \ref{fig:visualize_GaussianKernel}.

\subsubsection{Hide-and-Seek}\cite{singh2017hide} is a weakly supervised framework that has been proposed to improve object localization. Training object localization in a weakly supervised manner using intact images leads to poor localization by finding only the most salient parts of the objects. Hiding some random patches of the object during training helps to improve object localization by forcing the system to find relatively less salient parts. We have found that Hide-and-Seek can improve VOS which is a fully supervised learning task. As a result, we achieved comparable performance to the other offline-learning approaches, even when we trained only on the static images.

\subsubsection{Difficulties in segmentation near object boundaries.} 
Although there has been significant progress in image segmentation, accurate segmentation of the object boundary is still challenging. A low-level layer has been trained in EGNet \cite{zhao2019egnet} using object boundaries to accurately predict object boundaries. The imbalance between boundary pixels and non-boundary pixels has been addressed in LDF \cite{wei2020label} by separating them and training them separately. In this paper, we deal with the problem of GTs that are inaccurate near the object boundary. Hide-and-Seek addresses the problem by generating clean boundaries.

\begin{figure}[t]
\centering
\includegraphics[width=0.99\linewidth]{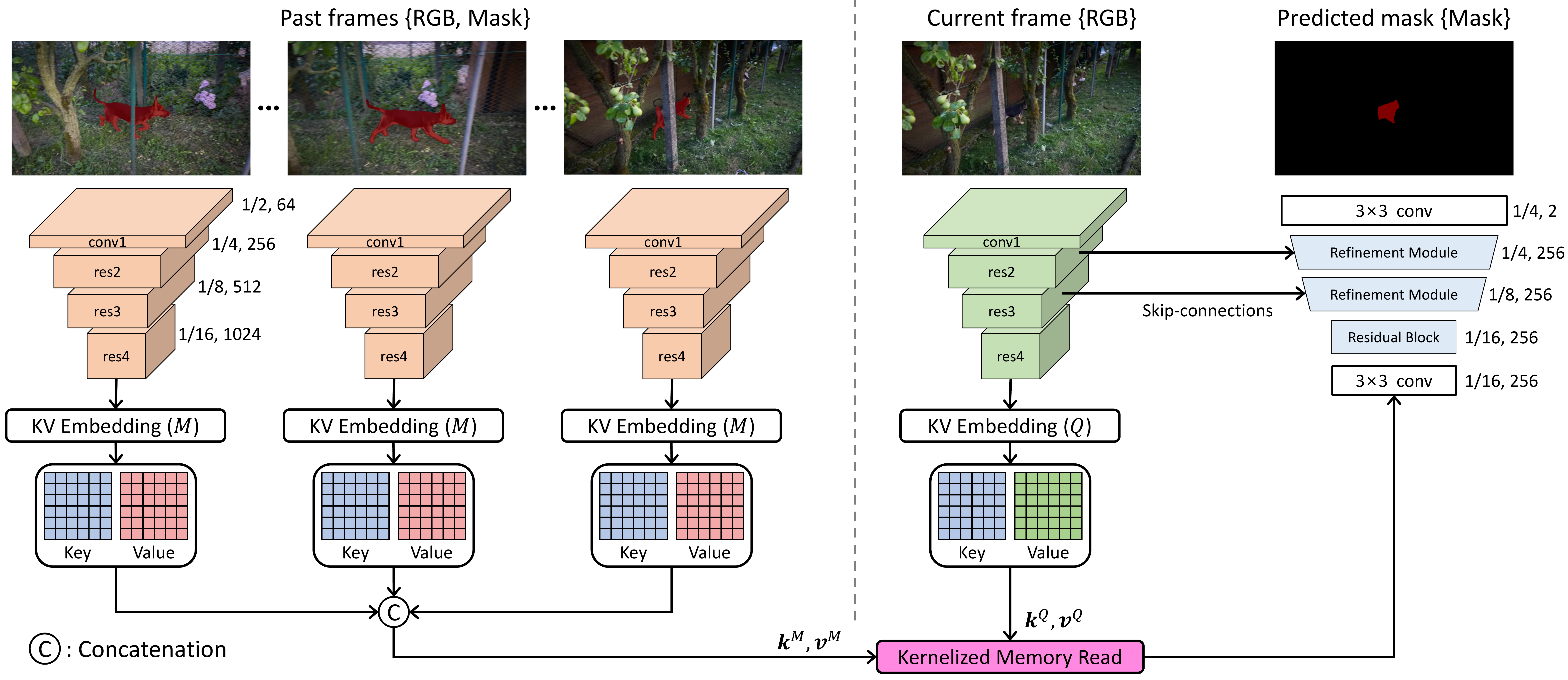}
\caption{
Overall architecture of our kernelized memory network (KMN). We follow the frameworks of \cite{Oh_2019_ICCV} and propose a new operation of kernelized memory read. The numbers next to the block indicate the spatial size and channel dimension, respectively.
}
\label{fig:overall_architecture}
\end{figure}

\section{Kernelized Memory Network}
\label{s3}

\subsection{Architecture}
\label{s31}
In this section, we present a kernelized memory network (KMN). The overall architecture of KMN is fairly similar to that of STM \cite{Oh_2019_ICCV}, as illustrated in Fig. \ref{fig:overall_architecture}. As in STM \cite{Oh_2019_ICCV}, the current frame is used as the query, while the past frames with the predicted masks are used as the memory. Two ResNet50 \cite{b23} are employed to extract the \textbf{key} and \textbf{value} from the memory and query frames. In memory, the predicted (or given) mask input is concatenated with the RGB channels. Then, the \textbf{key} and \textbf{value} features of the memory and the query are embedded via a convolutional layer from the \texttt{res4} feature \cite{b23}, which has a $1/16$ resolution resolution with respect to the input image. The structures of the \textbf{key} and \textbf{value} embedding layers for the query and memory are the same, but the weights are not shared. The memory may take several frames, and all frames in the memory are independently embedded and then concatenated along the temporal dimension. In the query, because it takes a single frame, the embedded \textbf{key} and \textbf{value} are directly used for memory reading.

The correlation map between the query and memory is generated by applying the inner product to all possible combinations of \textbf{key} features in the query and memory. From the correlation map, highly matched pixels are retrieved through a \textit{kernelized memory read} operation, and the corresponding \textbf{values} of the matched pixels in the memory are concatenated with the \textbf{value} of the query. Subsequently, the concatenated value tensor is fed to a decoder consisting of a residual block \cite{he2016identity} and two stacks of refinement modules. The refinement module is the same as that used in \cite{Oh_2019_ICCV,wug2018fast}. We recommend that the readers refer to \cite{Oh_2019_ICCV} for more details about the decoder.

The main innovation in KMN, distinct from STM \cite{Oh_2019_ICCV}, lies in the memory read operation. In the memory read of STM \cite{Oh_2019_ICCV}, only \texttt{Query-to-Memory} matching is conducted. In the kernelized memory read of KMN. however, both \texttt{Query-to-Memory} matching and \texttt{Memory-to-Query} matching are conducted. A detailed explanation of the kernelized memory read is provided in the next subsection.

\subsection{Kernelized Memory Read}
\label{s32}
In the memory read operation of STM \cite{Oh_2019_ICCV}, the non-local correlation map $c ( {\bf{p}} , {\bf{q}} )$ is generated using the embedded \textbf{key} of the memory ${{\bf{k}}^{M}} = \left\{ {{{k}}^{M}} ( {\bf{p}} ) \right\} \in {\mathbb{R}^{T \times H \times W \times C/8}}$ and query ${{\bf{k}}^{Q}} = \left\{ {{{k}}^{Q}} ( {\bf{q}} ) \right\} \in {\mathbb{R}^{H \times W \times C/8}}$ as follows:
\begin{equation}
c \left( {\bf{p}} , {\bf{q}} \right) = {{{{k}}^{M}}({\bf{p}})} {k^Q}({\bf{q}})^\top
\label{eq1}
\end{equation}
where $H$, $W$, and $C$ are the height, width, and channel size of \texttt{res4} \cite{b23}, respectively. ${\bf{p}} = \left[ p_t, p_y, p_x \right]$ and ${\bf{q}} = \left[ q_y , q_x \right]$ indicate the grid cell positions of the \textbf{key} features. Then, the query at position ${\bf{q}}$ retrieves the corresponding \textbf{value} from the memory using the correlation map by
\begin{equation}
r\left( {\bf{q}} \right) = \sum\limits_{\bf{p}} {\frac{{\exp \left( {c\left( {{\bf{p}},{\bf{q}}} \right)} \right)}}{{\sum\limits_{\bf{p}} {\exp } \left( {c\left( {{\bf{p}},{\bf{q}}} \right)} \right)}}} {v^M}\left( {\bf{p}} \right)
\label{eq2}
\end{equation}
where ${{\bf{v}}^{M}} = \left\{ {{{v}}^{M}} ( {\bf{p}} ) \right\} \in {\mathbb{R}^{T \times H \times W \times C/2}}$ is the embedded \textbf{value} of the memory. Then the retrieved \textbf{value} ${r}({\bf{q}})$, which is of size ${H \times W \times C/2}$, is concatenated with the query \textbf{value} ${{\bf{v}}^{Q}}\in {\mathbb{R}^{H \times W \times C/2}}$, and the concatenation result is fed to the decoder.

The memory read operation of STM \cite{Oh_2019_ICCV} has two inherent problems. First, every grid in the query frame searches the memory frames for a target object, but not vice versa. That is, there is only \texttt{Query-to-Memory} matching in the STM. Thus, when multiple objects in the query frame look like a target object, all of them can be matched with the same target object in the memory frames. Second, the non-local matching in the STM can be ineffective in VOS, because it overlooks the fact that the target object in the query should appear where it previously was in the memory frames.

\begin{figure}[t]
\centering
\begin{minipage}[]{0.5\linewidth}
\includegraphics[width=\linewidth]{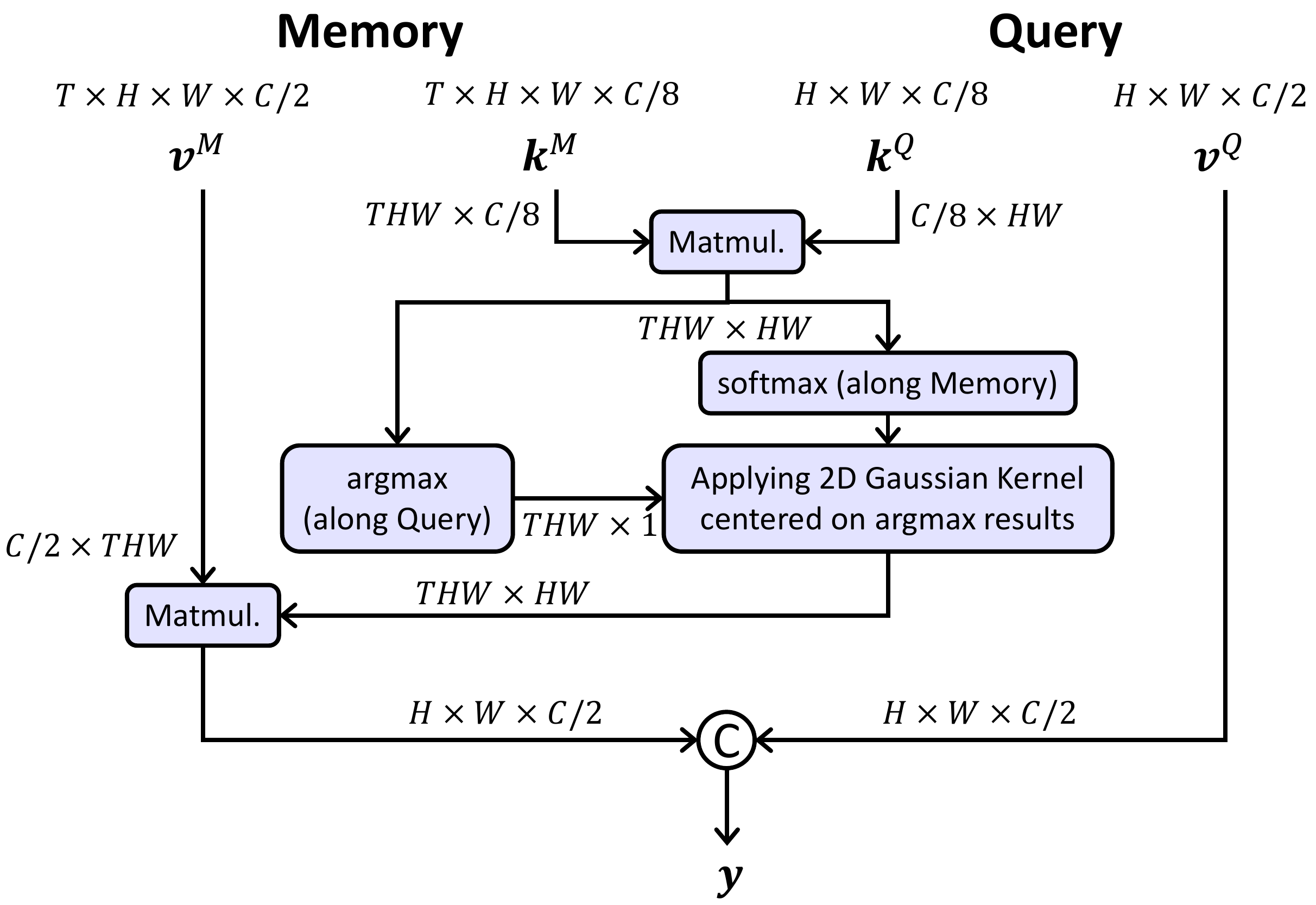}
\caption{Kernelized memory read operation.}
\label{fig:kernelized_memory_read}
\end{minipage}%
    \hfill%
\begin{minipage}[]{0.45\linewidth}
\includegraphics[width=\linewidth]{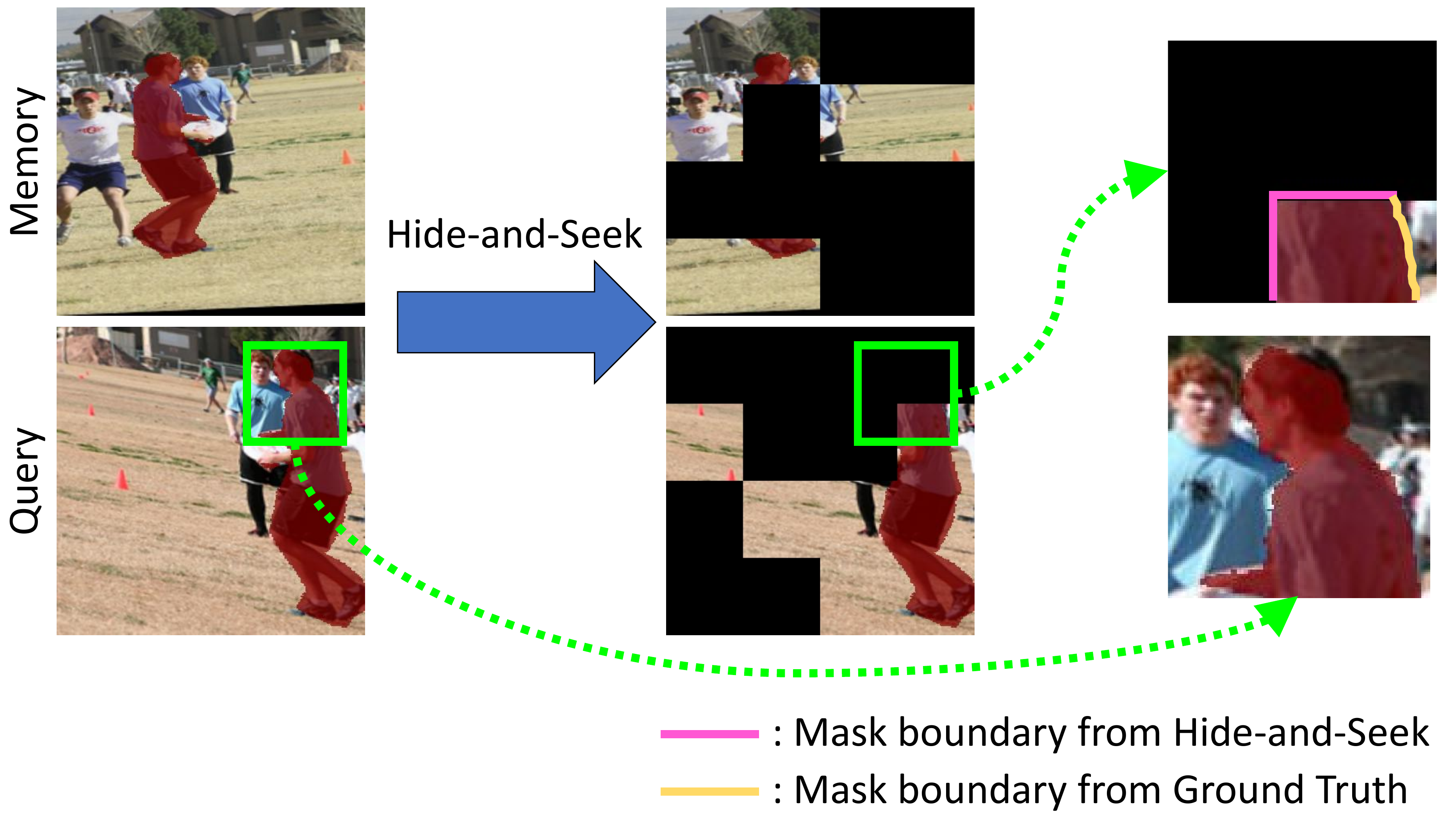}
\caption{A pair of images generated during pre-training using Hide-and-Seek. The mask indicated in red denotes the ground truth of the target object.}
\label{fig:hide-and-seek}
\end{minipage} 
\end{figure}

To solve these problems, we propose a kernelized memory read operation using 2D Gaussian kernels. First, the non-local correlation map $c \left( {\bf{p}} , {\bf{q}} \right) = {{{{k}}^{M}}({\bf{p}})} {k^Q}({\bf{q}})^\top$ between the query and memory is computed as in STM. Second, for each grid $\bf{p}$ in the memory frames, the best-matched query position ${\widehat{\bf{q}}}\left({\bf{p}}\right) = \left[ {{{\widehat q}_y}}\left({\bf{p}}\right), {{{\widehat q}_x}}\left({\bf{p}}\right) \right]$ is searched by
\begin{equation}
\widehat {\bf{q}}\left( {\bf{p}} \right) = \mathop {\arg\max }\limits_{\bf{q}} c\left( {{\bf{p}},{\bf{q}}} \right) .
\label{eq3}
\end{equation}
This is a \texttt{Memory-to-Query} matching. Third, a 2D Gaussian kernel ${\bf{g}} = \left\{g\left({\bf{p}} , {\bf{q}}\right)\right\} \in {\mathbb{R}^{T \times H \times W \times H \times W}}$ centered on $\widehat {\bf{q}}\left( {\bf{p}} \right) $ is computed by
\begin{equation}
g\left( {{\bf{p}},{\bf{q}}} \right) = \exp \left( { - \frac{{{{\left( {{q_y} - {{\widehat q}_y}\left( {\bf{p}} \right)} \right)}^2} + {{\left( {{q_x} - {{\widehat q}_x}\left( {\bf{p}} \right)} \right)}^2}}}{{2{\sigma ^2}}}} \right)
\label{eq4}
\end{equation}
where $\sigma$ is the standard deviation. Using Gaussian kernels, the \textbf{value} in the memory is retrieved in a local manner as follows:
\begin{equation}
{r^k}\left( {\bf{q}} \right) = \sum\limits_{\bf{p}} {\frac{{\exp \left( {c\left( {{\bf{p}},{\bf{q}}} \right)/\sqrt d } \right)g\left( {{\bf{p}},{\bf{q}}} \right)}}{{\sum\limits_{\bf{p}} {\exp \left( {c\left( {{\bf{p}},{\bf{q}}} \right)/\sqrt d } \right)g\left( {{\bf{p}},{\bf{q}}} \right)} }}} {v^M}\left( {\bf{p}} \right)
\label{eq5}
\end{equation}
where $d$ is the channel size of the \textbf{key}. This is a \texttt{Query-to-Memory} matching. Here, $\frac{1}{\sqrt{d}}$ is a scaling factor adopted from \cite{vaswani2017attention}, to prevent the argument in the softmax from becoming large in magnitude, or equivalently, to prevent the softmax from becoming saturated. The kernelized memory read operation is summarized in Fig. \ref{fig:kernelized_memory_read}.

\section{Pre-training by Hide-and-Seek}
\label{s33}
As in previous studies \cite{perazzi2017learning,wug2018fast,Oh_2019_ICCV}, our KMN is pre-trained using static image datasets that include foreground object masks \cite{b42,b43,hariharan2011semantic,shi2015hierarchical,cheng2014global,wang2017salient}. The basic idea of pre-training a VOS network is to synthetically generate a video with foreground object masks from a single static image. Applying random affine transforms to a static image and the corresponding object mask can yield a synthetic video, and the video can be used to pre-train a VOS network. The problem with synthetic generation of a video from a static image, however, is that the occlusion of the target object does not occur in a generated video. Thus, the simulated video cannot train the pre-trained KMN to cope with the common problem of occlusion in VOS. To solve this problem, the Hide-and-Seek strategy is used to synthetically generate a video with occlusions. Some patches are randomly hidden or blocked, and the occlusions are synthetically generated in the training samples. Here, we only consider squared occluders, but any shape can be taken. Hide-and-Seek can pre-train KMN to be robust to occlusion in the VOS. This idea is illustrated in Fig. \ref{fig:hide-and-seek}.

Further, it should be noted that most segmentation datasets contain inaccurate masks (GTs) near the object boundaries. Pre-training KMN with accurate masks is of great importance for high-performance VOS, because inaccurate masks can lead to performance degradation. Manual correction of incorrect masks would be helpful, but it would require a tremendous amount of labor. Another benefit obtained by the use of Hide-and-Seek in pre-training KMN is that the boundaries of the object segment become cleaner and more accurate than before. An example is illustrated in Fig. \ref{fig:hide-and-seek}. In this figure, the ground truth mask contains incorrect boundaries on the head of the running person. However, Hide-and-Seek creates a clear object boundary, as represented by the pink line in Fig. \ref{fig:hide-and-seek}. A detailed experimental analysis is given in Section \ref{s46}.

The use of Hide-and-Seek in the pre-training on simulated videos significantly improves the VOS pre-training performance; the results are given in Table \ref{tab_davis16}. The pre-training performance obtained by Hide-and-Seek is much higher than that of the previous methods \cite{wug2018fast,Oh_2019_ICCV}, and the performance is even as high as the full-training performance of some previous methods.

\section{Experiments}
\label{s4}
In this section, we describe the implementation details of the method, our experimental results on DAVIS 2016, DAVIS 2017, and Youtube-VOS 2018, and the analysis of our proposed methods.

\subsection{Training Details}
\label{s41}
We divide the training stage into two phases: one for pre-training on the static images and another for the main training on VOS datasets composed of video sequences.

During the pre-training, we generated three frames using a single static image by randomly applying rotation, flip, color jittering, and cropping, similar to \cite{wug2018fast,Oh_2019_ICCV}. We then used the Hide-and-Seek framework, as described in Section \ref{s33}. We first divided the image into a $24 \times 24$ grid, which has the same spatial size as the \textbf{key} feature. Each cell in the grid had a uniform probability to be hidden. We gradually increased the probability from 0 to 0.5.

During the main training, we followed the STM training strategy \cite{Oh_2019_ICCV}. We sampled the three frames from a single video. They were sampled in time-order with intervals randomly selected in the range of the maximum interval. In the training process, the maximum interval is gradually increased from 0 to 25.

For both training phases, we used the dynamic memory strategy \cite{Oh_2019_ICCV}. To deal with multi-object segmentation, a soft aggregation operation \cite{Oh_2019_ICCV} was used. Note that the Gaussian kernel was not applied during training. Because the argmax function, which determines the center point of the Gaussian kernel, is a discrete function, the error of the argmax cannot be propagated backward during training. Thus, if the Gaussian kernel is used during training, it attempts to optimize networks based on the incorrectly selected \textbf{key} feature by argmax, which leads to performance degradation.

Other training details are as follows: randomly resize and crop the images to the size of $384 \times 384$, use the mini-batch size of 4, minimize the cross-entropy loss for every pixel-level prediction, and opt for Adam optimizer \cite{KingmaB14} with a fixed learning rate of 1e-5.

\begin{table}
\caption{Comparisons on the DAVIS 2016 and DAVIS 2017 validation set where ground truths are available. `OL' indicates the use of online-learning strategy. The best results are \textbf{bold-faced}, and the second best results are \underline{underlined}.
}
\label{tab_davis16}
\centering
\begin{tabular}{clc|cccc|ccc}
\toprule
                               &                                        &            & \multicolumn{4}{c|}{DAVIS 2016 val}                                & \multicolumn{3}{c}{DAVIS 2017 val}                     \\
\multicolumn{1}{c}{Training Data} & \multicolumn{1}{c}{Methods}            & \; OL \;        & Time     & $\mathcal{G_M}$ & $\mathcal{J_M}$ & $\mathcal{F_M}$ & $\mathcal{G_M}$ & $\mathcal{J_M}$ & $\mathcal{F_M}$ \\
\midrule

\multirow{3}{*}{Static Images} & RGMP  \cite{wug2018fast}               &            & 0.13$s$  & 57.1            & 55.0            & 59.1            & -               & -               & -               \\
                               & STM \cite{Oh_2019_ICCV}                &            & 0.16$s$  & -               & -               & -               & 60.0            & 57.9            & 62.1            \\
                               & KMN (ours)                             &            & 0.12$s$  & \textbf{74.8}   & \textbf{74.7}   & \textbf{74.8}   & \textbf{68.9} & \textbf{67.1} & \textbf{70.8}               \\
\midrule
\multirow{30}{*}{DAVIS}       & BVS \cite{marki2016bilateral}          &            & 0.37$s$  & 59.4            & 60.0            & 58.8            & -               & -               & -               \\
                               & OSMN \cite{yang2018efficient}          &            & -        & -               & -               & -               & 54.8            & 52.5            & 57.1            \\
                               & OFL \cite{tsai2016video}               &            & 120$s$   & 65.7            & 68.0            & 63.4            & -               & -               & -               \\
                               & PLM \cite{shin2017pixel}               & \checkmark & 0.3$s$   & 66.0            & 70.0            & 62.0            & -               & -               & -               \\
                               & VPN \cite{jampani2017video}            &            & 0.63$s$  & 67.9            & 70.2            & 65.5            & -               & -               & -               \\
                               & OSMN \cite{yang2018efficient}          &            & 0.14$s$  & 73.5            & 74.0            & 72.9            & -               & -               & -               \\
                               & SFL \cite{cheng2017segflow}            & \checkmark & 7.9$s$   & 74.7            & 74.8            & 74.5            & -               & -               & -               \\
                               & PML \cite{chen2018blazingly}           &            & 0.27$s$  & 77.4            & 75.5            & 79.3            & -               & -               & -               \\
                               & MSK \cite{perazzi2017learning}         & \checkmark & 12$s$    & 77.6            & 79.7            & 75.4            & -               & -               & -               \\
                               & OSVOS \cite{caelles2017one}            & \checkmark & 9$s$     & 80.2            & 79.8            & 80.6            & 60.3            & 56.6            & 63.9            \\
                               & MaskRNN \cite{hu2017maskrnn}           & \checkmark & -        & 80.8            & 80.7            & 80.9            & -               & 60.5            & -               \\
                               & VidMatch \cite{hu2018videomatch}       &            & 0.32$s$  & -               & 81.0            & -               & 62.4            & 56.5            & 68.2            \\
                               & FAVOS \cite{cheng2018fast}             &            & 1.8$s$   & 81.0            & 82.4            & 79.5            & 58.2            & 54.6            & 61.8            \\
                               & LSE \cite{ci2018video}                 & \checkmark & -        & 81.6            & 82.9            & 80.3            & -               & -               & -               \\
                               & FEELVOS \cite{voigtlaender2019feelvos} &            & 0.45$s$  & 81.7            & 80.3            & 83.1            & 69.1            & 65.9            & 72.3            \\
                               & RGMP \cite{wug2018fast}                &            & 0.13$s$  & 81.8            & 81.5            & 82.0            & 66.7            & 64.8            & 68.6            \\
                               & DTN \cite{Zhang_2019_ICCV}             &            & 0.07$s$  & 83.6            & 83.7            & 83.5            & -               & -               & -               \\
                               & CINN \cite{bao2018cnn}                 & \checkmark & $>$30$s$ & 84.2            & 83.4            & 85.0            & 70.7            & 67.2            & 74.2            \\
                               & DyeNet \cite{li2018video}              &            & 0.42$s$  & -               & 84.7            & -               & 69.1            & 67.3            & 71.0            \\
                               & RaNet \cite{Wang_2019_ICCV}            &            & 0.03$s$  & 85.5            & 85.5            & 85.4            & 65.7            & 63.2            & 68.2            \\
                               & AGSS-VOS \cite{Lin_2019_ICCV}          &            & -        & -               & -               & -               & 66.6            & 63.4            & 69.8            \\
                               & DTN \cite{Zhang_2019_ICCV}             &            & -        & -               & -               & -               & 67.4            & 64.2            & 70.6            \\
                               & OnAVOS \cite{voigtlaender2017online}   & \checkmark & 13$s$    & 85.5            & 86.1            & 84.9            & 67.9            & 64.5            & 71.2            \\
                               & OSVOS$^S$ \cite{maninis2018video}      & \checkmark & 4.5$s$   & 86.0            & 85.6            & 86.4            & 68.0            & 64.7            & 71.3            \\
                               & DMM-Net \cite{Zeng_2019_ICCV}          &            & -        & -               & -               & -               & 70.7            & 68.1            & 73.3            \\
                               & STM \cite{Oh_2019_ICCV}                &            & 0.16$s$  & 86.5            & 84.8            & \underline{88.1}            & 71.6            & 69.2            & 74.0            \\
                               & PReMVOS \cite{luiten2018premvos}       & \checkmark & 32.8$s$  & 86.8            & 84.9            & \textbf{88.6}   & \textbf{77.8}   & \underline{73.9}            & \textbf{81.7}   \\
                               & DyeNet \cite{li2018video}              & \checkmark & 2.32$s$  & -               & 86.2            & -               & -               & -               & -               \\
                               & RaNet \cite{Wang_2019_ICCV}            & \checkmark & 4$s$     & \underline{87.1}            & \underline{86.6}            & 87.6            & -               & -               & -               \\
                               & KMN (ours)                             &            & 0.12$s$  & \textbf{87.6}   & \textbf{87.1}   & \underline{88.1}            & \underline{76.0}            & \textbf{74.2}   & \underline{77.8}            \\
\midrule
\multirow{6}{*}{+Youtube-VOS}  & S2S \cite{xu2018youtube}               & \checkmark & 9$s$     & -               & 79.1            & -               & -               & -               & -               \\
                               & AGSS-VOS \cite{Lin_2019_ICCV}          &            & -        & -               & -               & -               & 67.4            & 64.9            & 69.9            \\
                               & A-GAME \cite{johnander2019generative}  &            & 0.07$s$  & -               & 82.0            & -               & 70.0            & 67.2            & 72.7            \\
                               & FEELVOS \cite{voigtlaender2019feelvos} &            & 0.45$s$  & 81.7            & 81.1            & 82.2            & 72.0            & 69.1            & 74.0            \\
                               & STM \cite{Oh_2019_ICCV}                &            & 0.16$s$  & \underline{89.3}            & \underline{88.7}            & \underline{89.9}            & \underline{81.8}            & \underline{79.2}            & \underline{84.3}            \\
                               & KMN (ours)                             &            & 0.12$s$  & \textbf{90.5}   & \textbf{89.5}   & \textbf{91.5}   & \textbf{82.8}   & \textbf{80.0}   & \textbf{85.6}  \\
\bottomrule
\end{tabular}
\end{table}

\subsection{Inference Details}
\label{s42}
Our network utilizes intermediate frames to obtain rich information about the target objects. For the inputs of the memory, intermediate frames use the softmax output of the network directly, while the first frame uses the given ground truth mask. Even though we predict all the frames in a sequence, using all the past frames as memory is not only computationally inefficient but also requires considerable GPU memory. Therefore, we follow the memory management strategy described in \cite{Oh_2019_ICCV}. Both the first and previous frames are always used. The other intermediate frames are selected at five-frame intervals. Remainders are dropped.

We empirically set the fixed standard deviation $\sigma$ of the Gaussian kernel in (\ref{eq4}) to 7. We did not utilize any test time augmentation (\textit{e.g.}, multi-crop testing) or post-processing (\textit{e.g.}, CRF) and used the original image without any pre-processing (\textit{e.g.}, optical flow).

\begin{table}[t]
\caption{Comparisons on the DAVIS 2017 test-dev and Youtube-VOS 2018 validation sets where ground truths are unavailable. `OL' indicates the use of online-learning strategy. The best results are \textbf{bold-faced}, and the second best results are \underline{underlined}.
}
\label{tab_yv2018}
\centering
\begin{tabular}{lc|ccc|ccccc}
\toprule
\multicolumn{1}{c}{}                   &            & \multicolumn{3}{c|}{DAVIS17 test-dev}                   & \multicolumn{5}{c}{Youtube-VOS 2018 val}                                                        \\
\multicolumn{1}{c}{Methods}            & \; OL \;   & $\mathcal{G_M}$ & $\mathcal{J_M}$ & $\mathcal{F_M}$  & Overall          & $\mathcal{J_S}$  & $\mathcal{J_U}$  & $\mathcal{F_S}$  & $\mathcal{F_U}$  \\
\midrule
OSMN \cite{yang2018efficient}          &            & 39.3             & 33.7             & 44.9             & 51.2             & 60.0             & 40.6             & 60.1             & 44.0             \\
FAVOS \cite{cheng2018fast}             &            & 43.6             & 42.9             & 44.2             & -                & -                & -                & -                & -                \\
DMM-Net+ \cite{Zeng_2019_ICCV}         &            & -                & -                & -                & 51.7             & 58.3             & 41.6             & 60.7             & 46.3             \\
MSK \cite{perazzi2017learning}         & \checkmark & -                & -                & -                & 53.1             & 59.9             & 45.0             & 59.5             & 47.9             \\
OSVOS \cite{caelles2017one}            & \checkmark & 50.9             & 47.0             & 54.8             & 58.8             & 59.8             & 54.2             & 60.5             & 60.7             \\
CapsuleVOS \cite{Duarte_2019_ICCV}     &            & 51.3             & 47.4             & 55.2             & 62.3             & 67.3             & 53.7             & 68.1             & 59.9             \\
OnAVOS \cite{voigtlaender2017online}   & \checkmark & 52.8             & 49.9             & 55.7             & 55.2             & 60.1             & 46.6             & 62.7             & 51.4             \\
RGMP \cite{wug2018fast}                &            & 52.9             & 51.3             & 54.4             & 53.8             & 59.5             & 45.2             & -                & -                \\
RaNet \cite{Wang_2019_ICCV}            &            & 53.4             & 55.3             & 57.2             & -                & -                & -                & -                & -                \\
OSVOS$^S$ \cite{maninis2018video}      & \checkmark & 57.5             & 52.9             & 62.1             & -                & -                & -                & -                & -                \\
FEELVOS \cite{voigtlaender2019feelvos} &            & 57.8             & 55.1             & 60.4             & -                & -                & -                & -                & -                \\
RVOS \cite{ventura2019rvos}            &            & -                & -                & -                & 56.8             & 63.6             & 45.5             & 67.2             & 51.0             \\
DMM-Net+ \cite{Zeng_2019_ICCV}         & \checkmark & -                & -                & -                & 58.0             & 60.3             & 50.6             & 53.5             & 57.4             \\
S2S \cite{xu2018youtube}               & \checkmark & -                & -                & -                & 64.4             & 71.0             & 55.5             & 70.0             & 61.2             \\
A-GAME \cite{johnander2019generative}  &            & -                & -                & -                & 66.1             & 67.8             & 60.8             & -                & -                \\
AGSS-VOS \cite{Lin_2019_ICCV}          &            & -                & -                & -                & 71.3             & 71.3             & 65.5             & 75.2             & 73.1             \\
Lucid \cite{khoreva2019lucid}          & \checkmark & 66.7             & 63.4             & 69.9             & -                & -                & -                & -                & -                \\
CINN \cite{bao2018cnn}                 & \checkmark & 67.5             & 64.5             & 70.5             & -                & -                & -                & -                & -                \\
DyeNet \cite{li2018video}              & \checkmark & 68.2             & 65.8             & 70.5             & -                & -                & -                & -                & -                \\
PReMVOS \cite{luiten2018premvos}       & \checkmark & 71.6             & 67.5             & \underline{75.7} & -                & -                & -                & -                & -                \\
STM \cite{Oh_2019_ICCV}                &            & \underline{72.2} & \underline{69.3} & 75.2             & \underline{79.4} & \underline{79.7} & \underline{72.8} & \underline{84.2} & \underline{80.9} \\
KMN (ours)                             &            & \textbf{77.2}    & \textbf{74.1}    & \textbf{80.3}    & \textbf{81.4}    & \textbf{81.4}    & \textbf{75.3}    & \textbf{85.6}    & \textbf{83.3}   \\
\bottomrule
\end{tabular}
\end{table}

\subsection{DAVIS 2016 and 2017}
\label{s43}
DAVIS 2016 \cite{perazzi2016benchmark} is an object-level annotated dataset that contains 20 video sequences with a single target per video for validation. DAVIS 2017 \cite{pont20172017} is an instance-level annotated dataset that contains 30 video sequences with multiple targets per video for validation. Both DAVIS validation sets are most commonly used in VOS to validate proposed methods. We measure the official metrics: the mean of the region similarity $\mathcal{J_M}$, the contour accuracy $\mathcal{F_M}$, and their average value $\mathcal{G_M}$. We used a single parameter set that was trained on the DAVIS 2017 training dataset, which contains 60 video sequences, to evaluate our model on DAVIS 2016 and DAVIS 2017 for a fair comparison with previous works \cite{wug2018fast,yang2018efficient,Oh_2019_ICCV}. The experimental results on the DAVIS 2016 and 2017 validation sets are given in Table \ref{tab_davis16}. We report three different results for each training data.

The results of the training with only static images show a significant margin of improvement from previous studies. In addition, the performances of our proposed network trained on the static images show results comparable to those of the other approaches trained on DAVIS. This indicates that our Hide-and-Seek pre-training approach uses the static images effectively for VOS in training. STM \cite{Oh_2019_ICCV} trained on DAVIS showed weak performance compared with the online-learning methods. However, our approach achieves almost similar or even higher performance than the online-learning methods, along with a fast runtime. Finally, the results trained on an additional training dataset, Youtube-VOS, showed the best performance among all existing VOS approaches. Because the ground truths of the DAVIS validation set are accessible to every user, tuning on the dataset is relatively easy. Therefore, to show that a method actually works well in general, we evaluate our approaches on the DAVIS 2017 test-dev benchmark, where ground truths are unavailable, with results shown in Table \ref{tab_yv2018}. In DAVIS 2017 test-dev experiments, for a fair comparison, we resize the input frame to be 600p as in STM \cite{Oh_2019_ICCV}. We find that our approach surpasses the state-of-the-art method by a significant margin (+5\% $\mathcal{G_M}$ score).

\begin{figure}[t]
\centering
\includegraphics[width=0.99\textwidth]{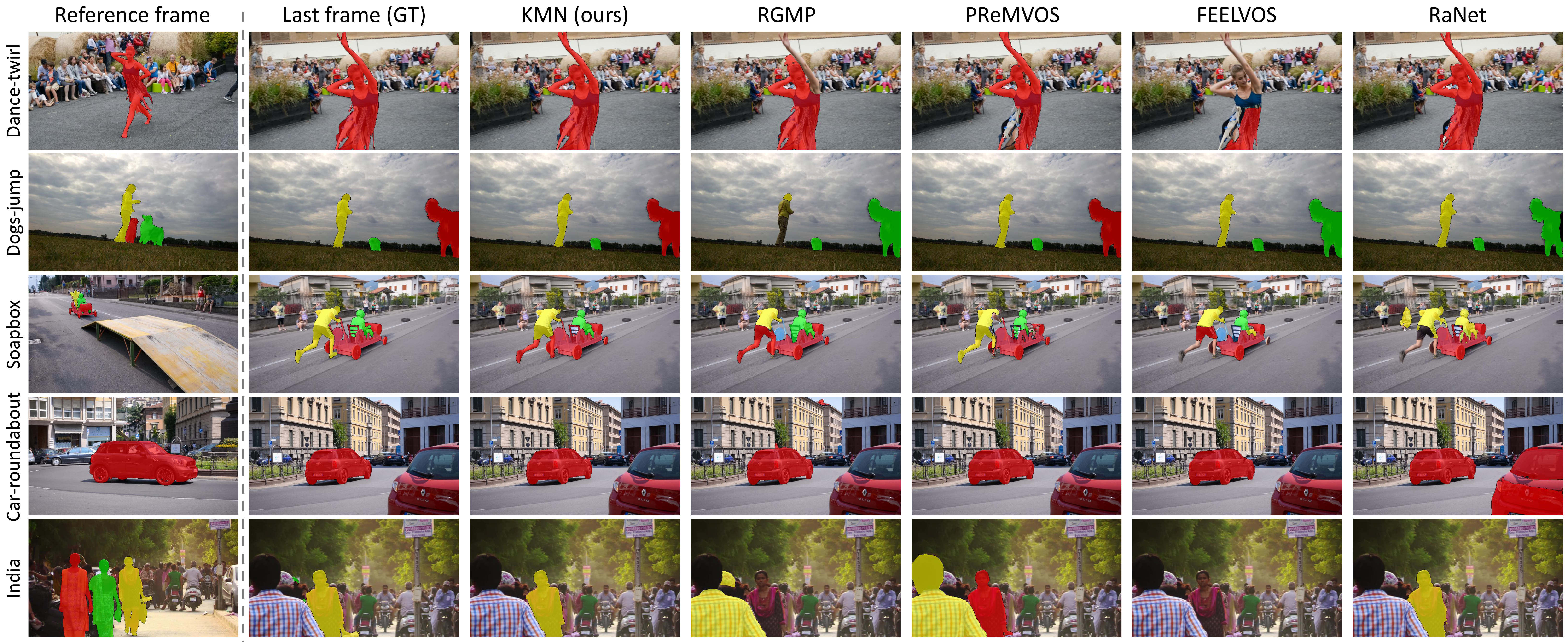}
\caption{
Qualitative results and comparisons on the DAVIS 2017 validation set. Our results also do not utilize additional training set, Youtube-VOS.
}
\label{fig:qualitative_results}
\end{figure}

\subsection{Youtube-VOS 2018}
\label{s44}
Youtube-VOS 2018 \cite{xu2018youtube} is the largest video object segmentation dataset. It contains 4,453 video sequences with multiple targets per video. To validate on Youtube-VOS 2018, both metrics $\mathcal{J}$ and $\mathcal{F}$ were calculated separately, depending on whether the object categories are seen or not during training: seen sequences with the number of 65 for $\mathcal{J_S}$, $\mathcal{F_S}$, and unseen sequences with the number of 26 for $\mathcal{J_U}$, $\mathcal{F_U}$. The ground truths of the Youtube-VOS 2018 validation set are unavailable as the DAVIS 2017 test-dev benchmark. As shown in Table \ref{tab_yv2018}, our approach achieved state-of-the-art performance. This indicates that our approach works well in all cases.

\begin{table}
\begin{threeparttable}
\caption{
Ablation study of our proposed methods. `HaS' and `KM' indicate the use of Hide-and-Seek pre-training and kernelized memory read operation, respectively. Note that we did not use additional VOS training data for the ablation study. Only either DAVIS or Youtube-VOS is used, depending on the target evaluation benchmark.
}
\label{tab_ablation}
\centering
\begin{tabular}{l|cc|cccc|ccc|ccccc}
\toprule
                      &            &            & \multicolumn{4}{c|}{DAVIS16}                                  & \multicolumn{3}{c|}{DAVIS17}                        & \multicolumn{5}{c}{Youtube-VOS 2018}                                                  \\
                       & HaS        & KM         & Time\tnote{$\star$}    & $\mathcal{G_M}$ & $\mathcal{J_M}$ & $\mathcal{F_M}$ & $\mathcal{G_M}$ & $\mathcal{J_M}$ & $\mathcal{F_M}$ & Overall       & $\mathcal{J_S}$ & $\mathcal{J_U}$ & $\mathcal{F_S}$ & $\mathcal{F_U}$ \\
\midrule
STM\cite{Oh_2019_ICCV} &            &            & 0.11$s$ & 86.5            & 84.8            & \textbf{88.1}   & 71.6            & 69.2            & 74.0            & 79.4          & 79.7            & 72.8            & 84.2            & 80.9            \\
\midrule
\multirow{4}{*}{Ours}  &            &            & 0.11$s$ & 81.3            & 80.0            & 82.6            & 72.6            & 70.1            & 75.0            & 79.0 & 79.2 & 73.5 & 83.1 & 80.3            \\
                       & \checkmark &            & 0.11$s$ & 87.1            & 86.3            & 88.0            & 75.9            & 73.7            & \textbf{78.1}   & 79.5          & 80.0            & 73.1            & 83.9            & 81.0            \\
                       &            & \checkmark & 0.12$s$ & 87.2            & 86.6            & 87.7            & 73.5            & 71.2            & 75.7            & 81.0          & 81.0            & \textbf{75.4}   & 85.0            & 82.5            \\
                       & \checkmark & \checkmark & 0.12$s$ & \textbf{87.6}   & \textbf{87.1}   & \textbf{88.1}   & \textbf{76.0}   & \textbf{74.2}   & 77.8            & \textbf{81.4} & \textbf{81.4}   & 75.3            & \textbf{85.6}   & \textbf{83.3}  \\
\bottomrule
\end{tabular}
{\small
\begin{tablenotes}
\item[$\star$] measured on our 1080Ti GPU system
\end{tablenotes}
}
\end{threeparttable}
\end{table}

\begin{figure}
\centering
\includegraphics[width=0.99\textwidth]{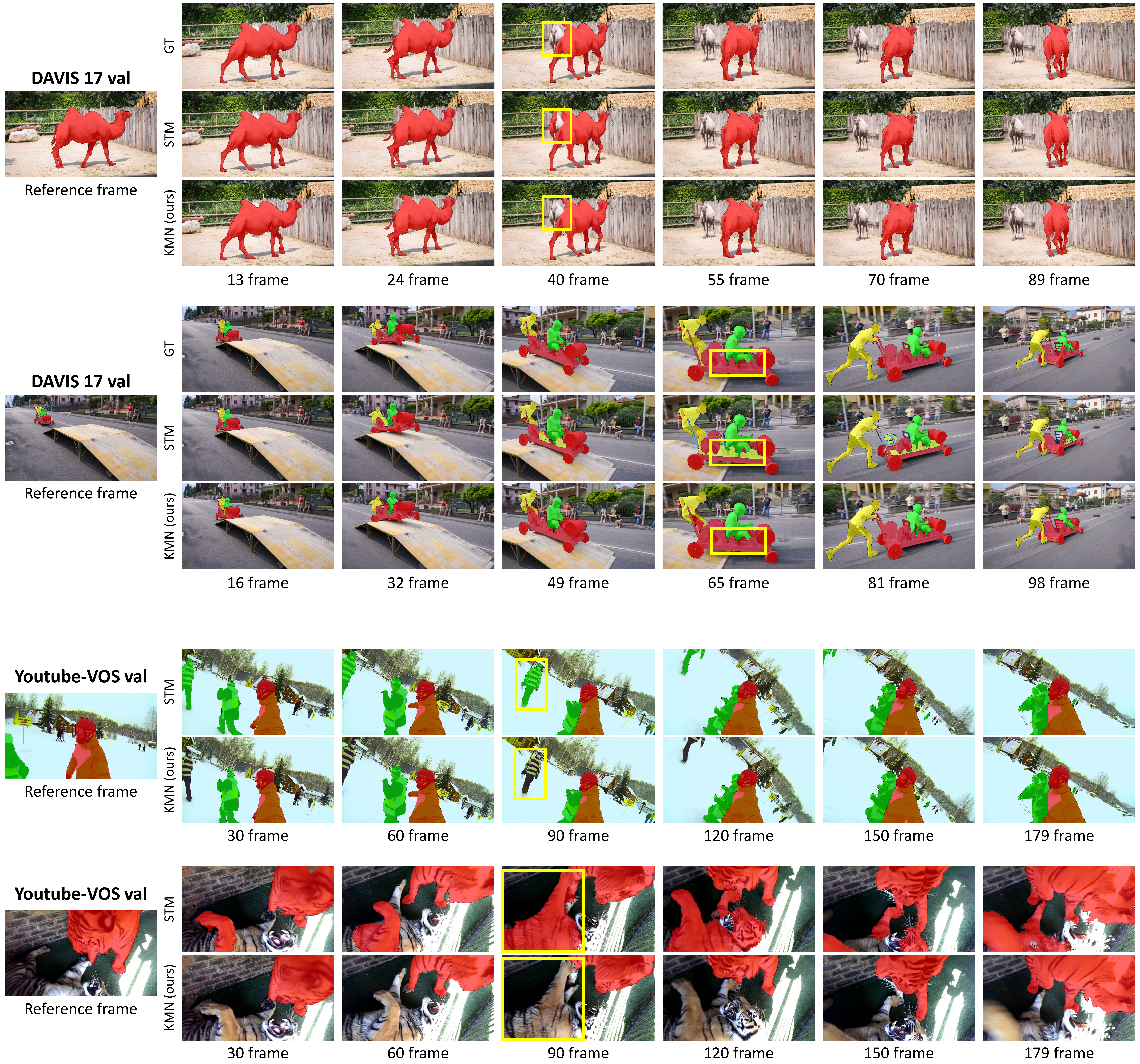}
\caption{
Qualitative results and comparisons with STM \cite{Oh_2019_ICCV}. The noticeable improvements are marked with yellow boxes. For DAVIS results, Youtube-VOS is additionally used for training. Note that the ground truths of the Youtube-VOS validation set are not available.
}
\label{fig:qualitative_results_STM}
\end{figure}

\subsection{Qualitative Results}
\label{s45}
A qualitative comparison is shown in Fig. \ref{fig:qualitative_results}. We compare our method with the state-of-the-art methods officially released on DAVIS\footnote{\url{https://davischallenge.org/davis2017/soa\_compare.html}}. The other methods in the figure do not utilize any additional VOS training data. Therefore, we show the KMN results which trained only on DAVIS in the main training stage for a fair comparison. Our results show consistently accurate predictions compared to other methods, even in cases of fast deformation (dance-twirl), the appearance of other objects, which are regarded as a background similar to the target object (car-roundabout), and the severe occlusion of the target objects (India).

\subsection{Analysis}
\label{s46}
\subsubsection{Ablation study.}
We conducted an ablation study to demonstrate the effectiveness of our approaches, and the experimental results are presented in Table \ref{tab_ablation}. As shown in the table, our approaches lead to performance improvements. The runtimes were measured on our 1080Ti GPU system, which is the same as that used in \cite{Oh_2019_ICCV}.

\begin{figure}[t]
\centering
\includegraphics[width=0.99\textwidth]{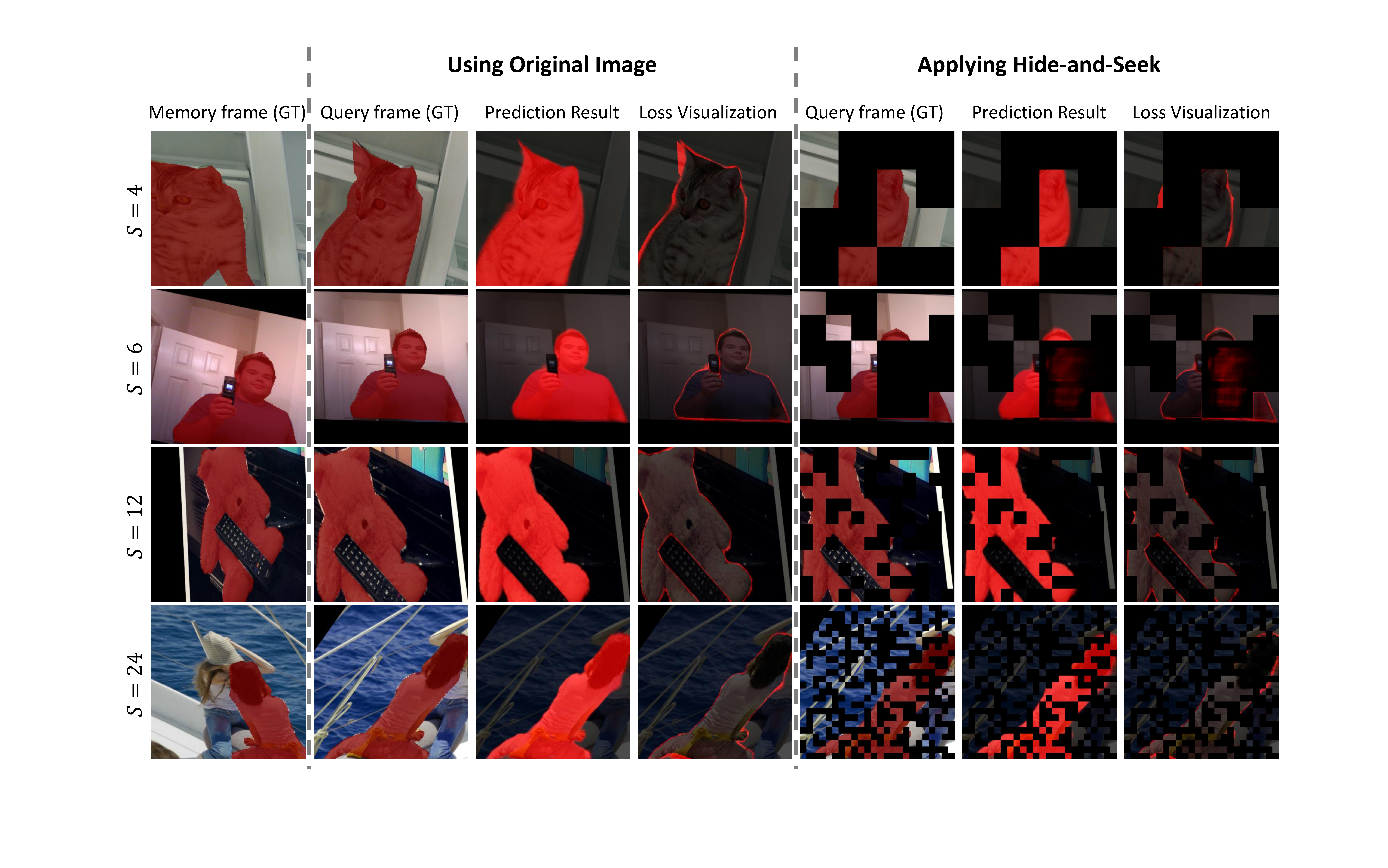}
\caption{
Pixel-level cross-entropy loss visualization during the pre-training on static images. `$S$' indicates the gird size of the Hide-and-Seek. Even if the network finds the object accurately, pixel-level losses occur near the mask boundary, because the ground truth masks near the boundary are not accurate. This makes it difficult for the network to learn the boundary correctly. Since  Hide-and-Seek can cut the object cleanly, it gives a more accurate ground truth mask near the boundary. Therefore, we can observe that the losses are not activated on the boundaries made by Hide-and-Seek.
}
\label{fig:loss_visualization}
\end{figure}

\subsubsection{Qualitative comparison with STM.}
We conducted a qualitative comparison with STM \cite{Oh_2019_ICCV}, and the results are shown in Fig. \ref{fig:qualitative_results_STM}. To show the actual improvements from STM, we obtained STM results using the author's officially released source code\footnote{\url{https://github.com/seoungwugoh/STM}}. However, since the parameters for Youtube-VOS validation are not available, our parameters shown in Table \ref{tab_ablation} were used for Youtube-VOS. For DAVIS, additional data, the Youtube-VOS set, was used for training. As shown in Fig. \ref{fig:qualitative_results_STM}, our results are robust and accurate even in difficult cases where \textit{multiple similar objects appear in the query} and \textit{occlusion occurs}.

\subsubsection{Boundary quality made by Hide-and-Seek.}
To verify that Hide-and-Seek modified the ground truth boundary accurately, we visualized the prediction loss for each pixel in Fig. \ref{fig:loss_visualization}. For a fair comparison, a single model trained on static images was used. As shown in the figure, \textit{most of the losses occur near the boundary}, even when the network predicts quite accurately. This indicates that the networks struggle to learn the mask boundary because the ground truth mask has an irregular and noisy boundary. However, \textit{the boundary of the hidden patch is not activated} in the figure. This means that the network can learn the mask boundary modified by Hide-and-Seek. Thus, Hide-and-Seek can provide more precise boundaries, and we expect that our new perspective would provide an opportunity to improve not only the quality of the segmentation masks, but also system performance for various segmentation tasks in the computer vision field.

\section{Conclusion}
\label{s5}
In this work, we present a new memory read operation and a method for handling occlusion and obtaining an accurate boundary using a static image. Our proposed methods were evaluated on the DAVIS 2016, DAVIS 2017, and Youtube-VOS benchmarks. We achieved state-of-the-art performance, even including online-learning methods. The ablation study shows the efficacy of our kernel approach, which addresses the main problem of memory networks in VOS. New approaches using the Hide-and-Seek strategy also show its effectiveness for VOS. Since our approaches can be easily reproduced and lead to significant improvements, we believe that our ideas have the potential to improve not only VOS, but also other segmentation-related fields.

\subsection*{Acknowledgement.}
This research was supported by Next-Generation Information Computing Development Program through the National Research Foundation of Korea(NRF) funded by the Ministry of Science, ICT (NRF-2017M3C4A7069370).

\clearpage
%
%
\bibliographystyle{splncs04}
\bibliography{main}
\end{document}